\documentclass[10pt,twocolumn,letterpaper]{article}

\usepackage{wacv}              

\usepackage{graphicx}
\usepackage{amsmath}
\usepackage{amssymb}
\usepackage{booktabs}
\usepackage[accsupp]{axessibility} 

\usepackage[pagebackref,breaklinks,colorlinks]{hyperref}

\usepackage[capitalize]{cleveref}
\crefname{section}{Sec.}{Secs.}
\Crefname{section}{Section}{Sections}
\Crefname{table}{Table}{Tables}
\crefname{table}{Tab.}{Tabs.}

\def\wacvPaperID{1151} 
\def\confName{WACV}
\def\confYear{2024}

\begin{document}

\title{Asymmetric Image Retrieval with Cross Model Compatible Ensembles}

\author{\textbf{Ori Linial}\thanks{Contributed equally.}\hspace{1.4mm}$^2$ \qquad \textbf{Alon Shoshan}\protect\footnotemark[1]\hspace{1.4mm}$^1$ \qquad \textbf{Nadav Bhonker}$^1$ \qquad \textbf{Elad Hirsch}$^2$ \\ 
\textbf{Lior Zamir}$^1$ \qquad \textbf{Igor Kviatkovsky}$^1$ \qquad \textbf{G\'{e}rard Medioni}$^1$ \\
$^1$Amazon \qquad
$^2$Technion - Israel Institute of Technology
}
\maketitle

\begin{abstract}
The asymmetrical retrieval setting is a well suited solution for resource constrained applications such as face recognition and image retrieval.
In this setting, a large model is used for indexing the gallery while a lightweight model is used for querying.
The key principle in such systems is ensuring that both models share the same embedding space.
Most methods in this domain are based on knowledge distillation.
While useful, they suffer from several drawbacks: they are upper-bounded by the performance of the single best model found and cannot be extended to use an ensemble of models in a straightforward manner.
In this paper we present an approach that does not rely on knowledge distillation, rather it utilizes embedding transformation models.
This allows the use of $N$ independently trained and diverse gallery models (e.g., trained on different datasets or having a different architecture) and a single query model.
As a result, we improve the overall accuracy beyond that of any single model while maintaining a low computational budget for querying.
Additionally, we propose a gallery image rejection method that utilizes the diversity between multiple transformed embeddings to estimate the uncertainty of gallery images.
\end{abstract}

\section{Introduction}
\label{sec:introduction}
\begin{figure}[t]
    \centering
    \begin{subfigure}[b]{1\linewidth}
        \includegraphics[width=0.98\linewidth]{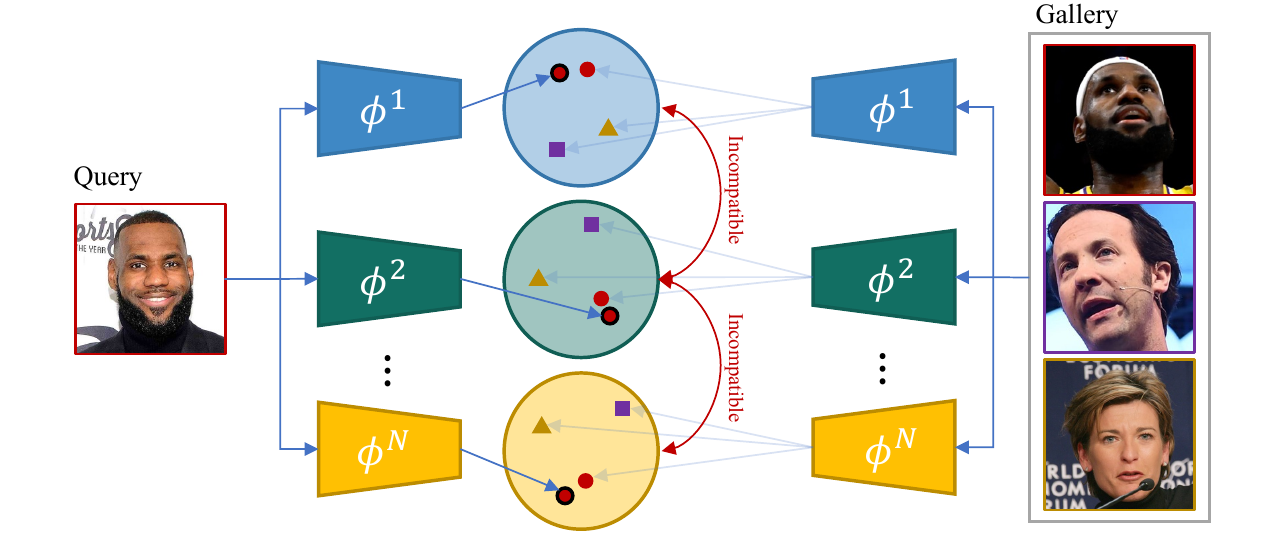}
        \vskip -0.05in
        \caption{Standard ensemble (symmetric).} 
        \label{fig:intro_a}
    \end{subfigure}
    \begin{subfigure}[b]{1\linewidth}
        \includegraphics[width=0.98\linewidth]{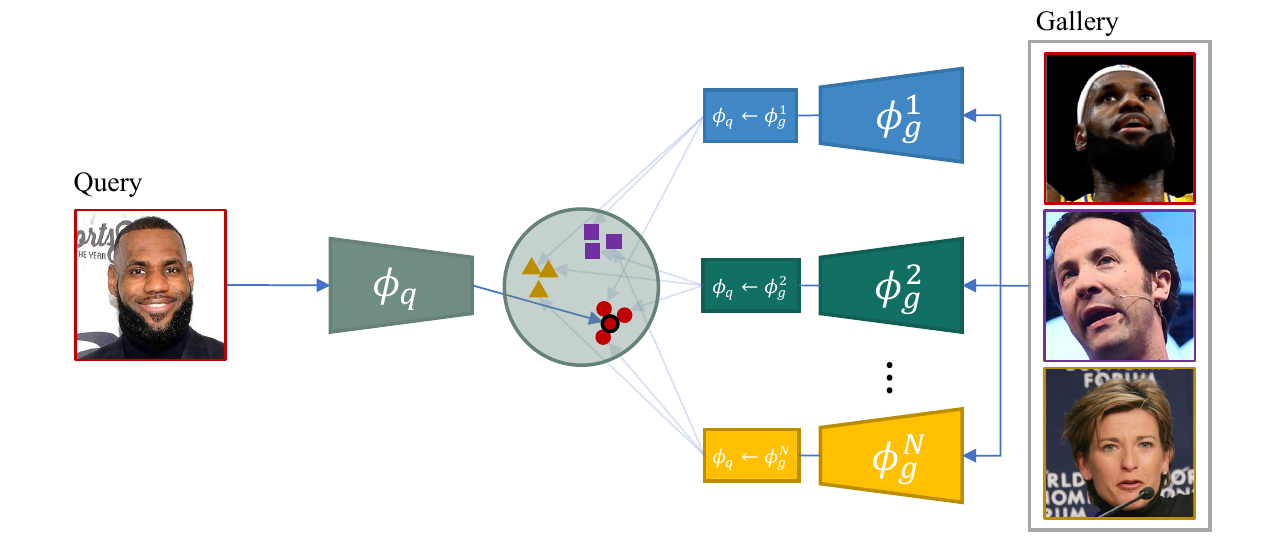}
        \vskip -0.05in
        \caption{Embedding transformation based ensemble (asymmetric).}
        \label{fig:intro_b}
    \end{subfigure}
    \caption{(a) Construction of an ensemble in the standard \textit{symmetric retrieval} setting requires that each model will be used both for computing the gallery and for querying, since the embedding spaces of all models are incompatible.
    Hence, user-side querying introduces heavy computational costs, requiring to compute embeddings of $N$ models and calculate distances in $N$ embedding spaces. 
    (b) We propose to construct an ensemble in an \textit{asymmetric retrieval} setting, where a single lightweight model is used for querying and an ensemble of diverse independently trained models is used for computing the gallery (a one time process that occurs offline).
    Using embedding transformation models, $\phi^i_g \rightarrow \phi_q$, all gallery models' embeddings are transformed to the embedding space of the query model.
    Our approach benefits both from the increased accuracy that is associated with ensembles and from low computational requirements for querying.
    }
    \vskip -0.12in
    \label{fig:intro}
\end{figure}


Face recognition and image retrieval at scale are among the most challenging and widely studied topics in computer vision, having many practical applications.
Modern systems are expected to provide extremely high retrieval accuracy under real time performance constraints and to support a very large number of classes (identities) in the gallery set. 
The majority of existing face recognition and image retrieval methods~\cite{babenko2014neural, radenovic2018fine, simeoni2019local, tolias2020learning, deng2019arcface, wang2017normface} utilize a \textit{symmetric retrieval} approach, \ie the same model is used for extracting feature vectors (embeddings) for the gallery and for the query images.
In a symmetric retrieval setting there is a clear trade-off between retrieval accuracy and computing resources. 
Using a larger model almost always achieves higher accuracy. 
However, it may prohibit use on user-side-devices due to limited computational and memory resources.
On the other hand, using a lightweight model usually results in inferior representation capability and lower overall retrieval accuracy.
Instead of compromising accuracy to meet hardware requirements, an \textit{asymmetric retrieval} setting~\cite{duggal2021compatibility, Wu_2022_CVPR, Budnik_2021_CVPR} can be incorporated, where a large model is used for indexing the gallery offline, with sufficient computational resources, and a lightweight model is used for processing query images.
The former is referred to as the \textit{gallery model} and latter as the \textit{query model}.  
Asymmetric retrieval is not straightforwardly accomplished by training both gallery and query models independently since their embedding spaces are incompatible.
This is also known as the \textit{cross model compatibility} (CMC) problem~\cite{wang2020unified}.

Previous asymmetric retrieval methods~\cite{duggal2021compatibility, Wu_2022_CVPR, Budnik_2021_CVPR} use knowledge distillation~\cite{hinton2015distilling} based approaches in order to restrict the query model embedding space to be compatible with that of the gallery model.
These methods show significant accuracy improvements compared to methods using only the lightweight query model both for indexing and for querying (symmetric retrieval). 
However, the accuracy of all these methods is also upper-bounded by the gallery model accuracy and is usually lower than when a large model is used for both indexing and querying. 
In this work we present a new asymmetric retrieval approach that is not based on knowledge distillation.
Instead, the approach utilizes embedding space transformations~\cite{wang2020unified} which allows to transform multiple embedding spaces of different gallery models into the one of the query model. 
In contrast to previous methods, the resulting accuracy is no longer upper-bounded by that of the gallery model.
We empirically demonstrate that our approach indeed breaches this ``upper-bound''.
We note that applying an ensemble of models in the case of knowledge distillation is not straightforward. 
For example, it is not possible to independently train all of the ensemble components as they all have to rely on a common predefined model. 
This means that models acquired independently from different sources\footnote{For example, if the models were trained on different datasets that may no longer be available.}, cannot be combined into an ensemble under such a distillation scheme. 
Additionally, distilling models with the same level of accuracy while optimizing for CMC was shown to reduce the accuracy of the resulting model~\cite{shen2020towards, wang2020unified}.



The symmetric setting implicitly assumes that using an ensemble of models for computing the gallery will increase the computational cost of user-side querying (since the same ensemble has to be used for querying).
It is not straightforward how to reduce the computational cost of the querying part since the model embedding spaces are incompatible, see Fig.~\ref{fig:intro}.
Attempts to remediate the incompatibility issue lead to asymmetric knowledge distillation methods, which, as described above, are sub-optimal for creating ensembles. 

In addition to improving asymmetric retrieval accuracy, we introduce an uncertainty based gallery image rejection method.
By leveraging the diversity between the multiple transformed embeddings of the same gallery image, we can estimate the uncertainty of the final transformed embedding.
For both the face recognition and product retrieval domain, we show that excluding embeddings with high uncertainty significantly improves the overall accuracy.
For example, by rejecting only 10\% of the gallery embeddings we reduce the face recognition error by 17.4\%.
This rejection approach is only possible because our method allows multiple models to project the same image into a common embedding space, which is not possible in a symmetric setting.

Our approach is also suitable in the \textit{compatible model update} setting~\cite{jaeckle2023fastfill, shen2020towards, wang2020unified, chen2019r3, zhou2022bt}.
In this setting, when a better model becomes available (by improved training data, model architectures, or training regimes), we are tasked to update the retrieval system using the new model without performing \textit{backfilling}.
Backfilling is the common na\"ive process of replacing the embeddings in the gallery set that have been generated by the old model with embeddings from the new model.
This method is computationally expensive and in some practical cases not even possible since the gallery images might not be retained by the system.

We summarize our contributions as following:
\begin{enumerate}
    \item We present a novel embedding transformation based ensemble for asymmetric image retrieval, which significantly improves accuracy without the need for additional computation when performing user-side querying.
    \item We introduce a gallery image rejection method that leverages the diversity between multiple transformed embeddings.
    This method can prevent ``hard to match'' gallery images from registering, thus further improving accuracy.
    \item We demonstrate that our approach can achieve high accuracy even in a challenging compatible model update setting.
\end{enumerate}

\section{Related work}
\label{sec:related_work}
\subsection{Image retrieval}
Modern image retrieval methods~\cite{razavian2016visual, zheng2015query, xie2015image, uijlings2013selective, babenko2015aggregating}, rely on deep learning models that encode images to a low dimensional embedding space.
Embedding vectors representing similar classes are mapped close to each other, while dissimilar ones are mapped far apart.
Such an embedding space can be created by using a classification loss as a proxy ~\cite{deng2019arcface,taigman2014deepface,WangCLL18,wang2017normface,SunCZZZWW20} or by using metric learning techniques such as triplet loss or contrastive loss ~\cite{hadsell2006dimensionality,sun2015deep,schroff2015facenet,HofferA15,BromleyBBGLMSS93,ChopraHL05}.
We note that all these approaches focus on the symmetrical retrieval setting. 

\subsection{Cross model compatibility}
In recent years, increased attention has been given to the cross model compatibility (CMC)~\cite{wang2020unified} problem.
The aim of this field is to ensure embeddings encoded by different models are compatible, a condition that is usually not possible when models are trained independently.
Compatibility between models is critical for asymmetric retrieval settings where the query and gallery models are different.
In a resource constrained scenario recent methods~\cite{duggal2021compatibility, Wu_2022_CVPR, Budnik_2021_CVPR} propose a large model for indexing and a lightweight model for user-side querying.
Compatible model update~\cite{shen2020towards, zhang2022towards, meng2021learning, chen2019r3, wang2020unified} is another asymmetric retrieval scenario where a new query model replaces an older one and backfilling is to be avoided (due to high computational cost or unavailability of gallery images).
The query model is usually better than the gallery model in this scenario since it was trained later with improved architecture, an improved training scheme or more training data.
Knowledge distillation based methods~\cite{duggal2021compatibility, Wu_2022_CVPR, Budnik_2021_CVPR, shen2020towards, zhang2022towards, meng2021learning} use a compatibility loss term during training, enforcing the query model's embedding space to reside in the same space as the gallery model's.
Chen~\etal~\cite{chen2019r3} propose $R^3AN$ that combines reconstruction, representation and regression techniques to transform embeddings from one model to another.
Wang~\etal~\cite{wang2020unified} propose to learn transformations from the embedding spaces of both trained models to a unified embedding space while enforcing compatibility.
In this work we demonstrate that our transformation based ensemble approach is applicable for both asymmetric retrieval scenarios.

\subsection{Using Ensembles}
Ensembles of models have been extensively used in machine learning to boost accuracy~\cite{DBLP:conf/nips/KrizhevskySH12,DBLP:journals/corr/SimonyanZ14a,zhang2012ensemble,sagi2018ensemble,kuncheva2014combining,dong2020survey,li2018research}. 
Diversity among the ensemble components is important for ensuring a performance gain compared to relying on each component individually~\cite{krogh1994neural,hansen1990neural}. 
A straightforward way to achieve such diversity is by introducing variations in training data~\cite{breiman1996bagging} or initialization conditions~\cite{kolen1990back}. 

An important aspect is designing the fusion scheme used to aggregate the ensemble component outputs~\cite{granitto2005neural}.
Such fusion may be performed in different stages of the ensemble.
For example, in a closed-set scenario the fusion is done by combining the softmax class posteriors, leveraging the common representation between models~\cite{simonyan2014very,szegedy2015going,he2016deep}.
Other approaches delay the fusion to an even later stage of combining final model predictions by performing weighted averaging or majority voting, \etc~\cite{ju2018relative}.
In this work we propose to use feature based fusion for image retrieval, combining embeddings of different models.
This approach is non-trivial since embedding spaces of different models are typically incompatible.
Despite the fact that ensembles improve accuracy significantly, they are often ignored due to their high resource requirements.
In the asymmetric retrieval settings this limitation is entirely avoided, thus making ensembles a well suited approach.

\section{Proposed approach}


\subsection{Image retrieval and cross model compatibility}
In a typical image retrieval system a set of \emph{gallery} images, $I_g$, are associated with $C$ classes (or identities), $Y_g=\{y_i\}_{i=1}^C$.
A gallery model, $\phi_g$, maps each image $i_g \in I_g$ to the gallery embedding space, $\mathcal{G}\subseteq\mathbb{R}^n$.
By applying $\phi_g$ to the entire gallery, we obtain the set of gallery embedding vectors $E_g$.
At test time, we are presented with a query image, $i_q$, belonging to some class, $y_q$ (not necessarily in $Y_g$).
The query image is then consumed by a query model, $\phi_q$, to produce a query embedding, $e_q$.
Assuming a symmetric setting ($\phi_g=\phi_q$), we associate $e_q$ with the class $Y_g$ of the closest gallery embedding vector based on some distance metric $d(\cdot,\cdot)$.
If $e_q$ has no sufficiently close match in $E_g$, the query image is rejected.
The CMC problem becomes relevant in an asymmetric setting ($\phi_g\neq\phi_q$), where both models are trained independently.
In this case, $\phi_q$ maps images into a query embedding space, $\mathcal{Q}\subseteq\mathbb{R}^m$, that is incompatible with the gallery embedding space, $\mathcal{G}$.

\subsection{Embedding transformation}
\subsubsection{Unified embedding space} 
To address the CMC problem, Wang~\etal~\cite{wang2020unified} suggest to train embedding transformation models, $T_g$ and $T_q$, to transform both $\phi_g$ and $\phi_q$ embedding spaces into a unified embedding space.
In the unified space, embedding vectors transformed from $\phi_g$ and $\phi_q$ are compatible with one another. 
To implement the transformation models, the authors propose using four consecutive Residual Bottleneck Transformation (RBT) modules~\cite{wang2020unified}, and a compatibility constraining training scheme.
The general training scheme is as follows: all training images are encoded using $\phi_g$ and $\phi_q$ to produce training embedding sets $F_g$ and $F_q$.
During training, corresponding embeddings from $F_g$ and $F_q$ are transformed by $T_g$ and $T_q$, respectively, to the unified space.
To enforce compatibility in the unified space, a combination of three loss terms are applied: a similarity-, a KL-divergence- and a dual-classification-loss.
The first two terms enforce similarity between embeddings, while the third term enforces the embedding spaces to be discriminative by identity.
Furthermore, by using a shared classification head, the embedding spaces are constrained to be aligned.




\subsubsection{Model-to-model transformation}
\label{sec:m2m}

In this work we modify the unified embedding approach so that only one embedding space is transformed.
Specifically, the gallery's embedding space is transformed to the query's.
This is achieved by following the same training scheme except that we set $T_q$ to be an identity mapping, \ie, we learn a model to model transformation (M2M).
Since the embedding space used for querying does not undergo a transformation, we gain the following practical benefits:
\begin{enumerate}
\item Multiple transformations from different embedding models' spaces to the same query embedding space can be learned using the same M2M training scheme. 
\item No additional parameters are added to the user side querying system, preserving its computational resource efficiency.
\end{enumerate}


\subsection{Cross model compatible ensembles}
Leveraging the above benefits and taking into account that gallery indexing is performed offline, we propose to register an image with multiple gallery models (Fig.~\ref{fig:intro_b}).
We train a set of $N$ gallery models, $\{\phi_g^i\}_{i=1}^N$, and a corresponding set of transformation models $\{T_g^i\}_{i=1}^N$.
During indexing, each gallery image is processed by all gallery models $\{\phi_g^i\}_{i=1}^N$, producing a set of embeddings $\{e^i_g\}_{i=1}^N$ for each image\footnote{In the compatible model update scenario, the gallery images can be discarded at this point.}.
We then apply $\{T_g^i\}_{i=1}^N$ to transform all embeddings to the embedding space of $\phi_q$.
Subsequently, we produce a single gallery embedding in $\phi_q$'s embedding space by averaging the transformed embeddings of each image:
\begin{align}
    \hat{e}_q = \frac{1}{N} \sum_{i=1}^{N} T_i (e_g^i).
\end{align}
where $\hat{e}_q$ is the final embedding of a single gallery image.
We emphasize that this proposition does not increase the test-time latency, query model size and the number of comparisons during the querying process.
Alternative approaches for combining multiple embedding spaces were considered.
In this section we presented the best performing approach.
Experiments of alternative approaches are presented in Section~\ref{sec:alternative}.

\begin{figure}
    \centering
    \includegraphics[width=0.45\textwidth]{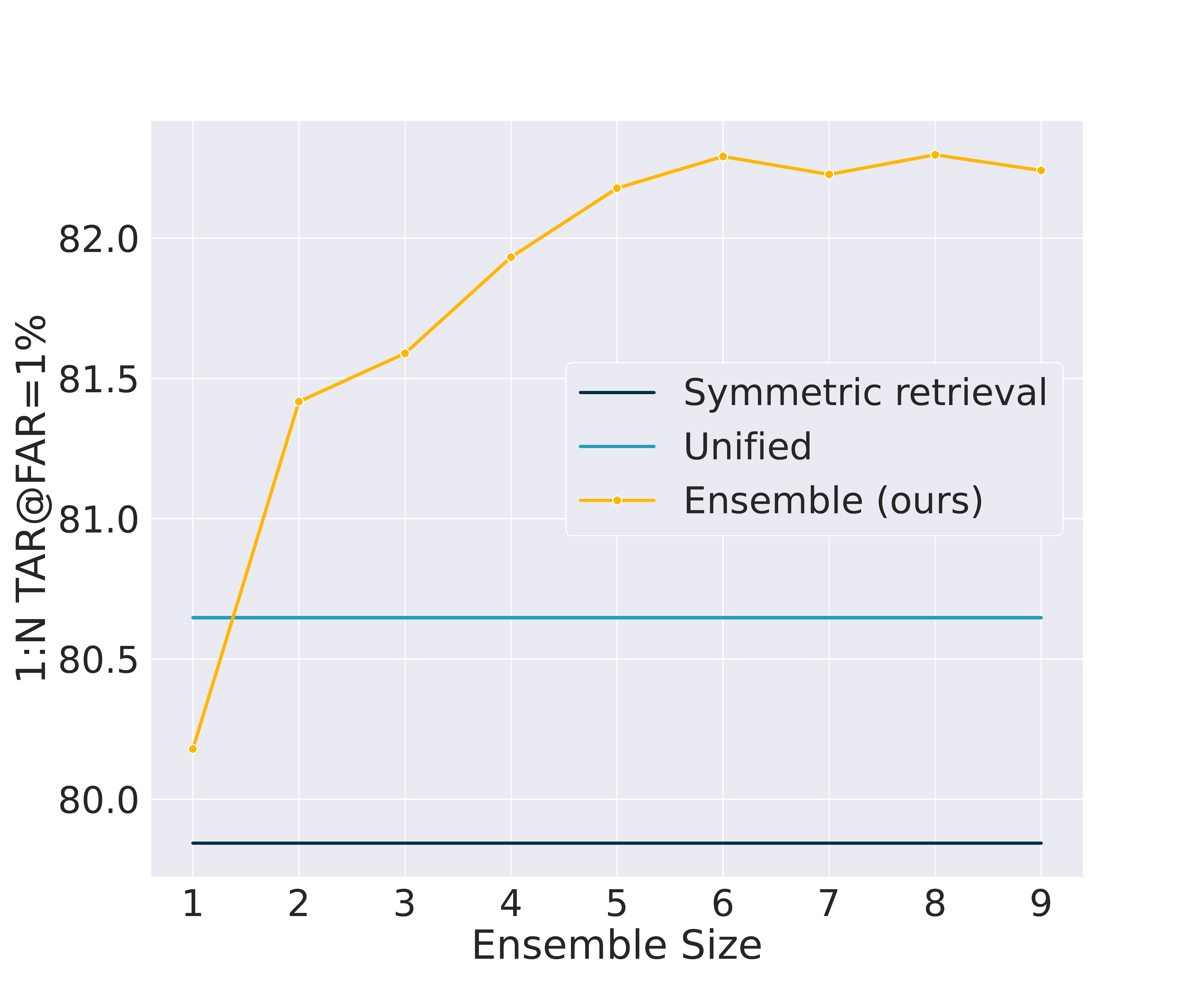}
    \label{fig:basline_comp}
    \caption{1:N search for increasing ensemble sizes.
    Since the symmetric retrieval setting and Unified use only one gallery model they are visualized as lines.
    }
    \label{fig:basline_comp}
\end{figure}
\begin{table}
    \centering
    \small{
    \begin{tabular}{l c c}
        \toprule
        \quad\quad\quad\quad\quad\quad\quad\quad\quad\quad\quad\quad & 1:N TAR & 1:1 TAR\\
        {} & @FAR=1\% & @FAR=0.01\% \\
        \midrule
        Symmetric retrieval & 79.843\tiny{$\pm 0.50$} & 88.064\tiny{$\pm 0.27$}\\
        \midrule
        Unified~\cite{wang2020unified} & 80.647\tiny{$\pm 0.32$}  & 88.439\tiny{$\pm 0.14$} \\
        M2M &  80.476\tiny{$\pm0.29$}  & 88.368\tiny{$\pm0.12$} \\
        \midrule
        Ensemble of 2 & 81.417 & 89.068 \\
        Ensemble of 4 & 81.932 & 89.139 \\
        Ensemble of 9 & \textbf{82.241} & \textbf{89.344} \\
        \bottomrule
    \end{tabular}
    }
    \caption{1:N search and 1:1 verification TAR (in percentage) at fixed FAR values.
    Symmetric retrieval, $\phi_g=\phi_q$, was evaluated on ten ResNet18 models.
    Unified and M2M were evaluated on ten different $\phi_g$ ResNet18 models and the same $\phi_q$ ResNet18 model in each evaluation.
    Symmetric retrieval, unified and M2M TARs are shown as mean and std values of ten evaluations.
    Our method was evaluated using the same $\phi_q$ model as Unified and M2M and ensemble sizes of 2, 4 and 9 for comupting the gallery. 
    }
    \label{tab:basline_comp}
\end{table}
\begin{figure*}
     \centering
     \begin{subfigure}[b]{0.241\textwidth}
         \centering
         \includegraphics[width=0.88\textwidth]{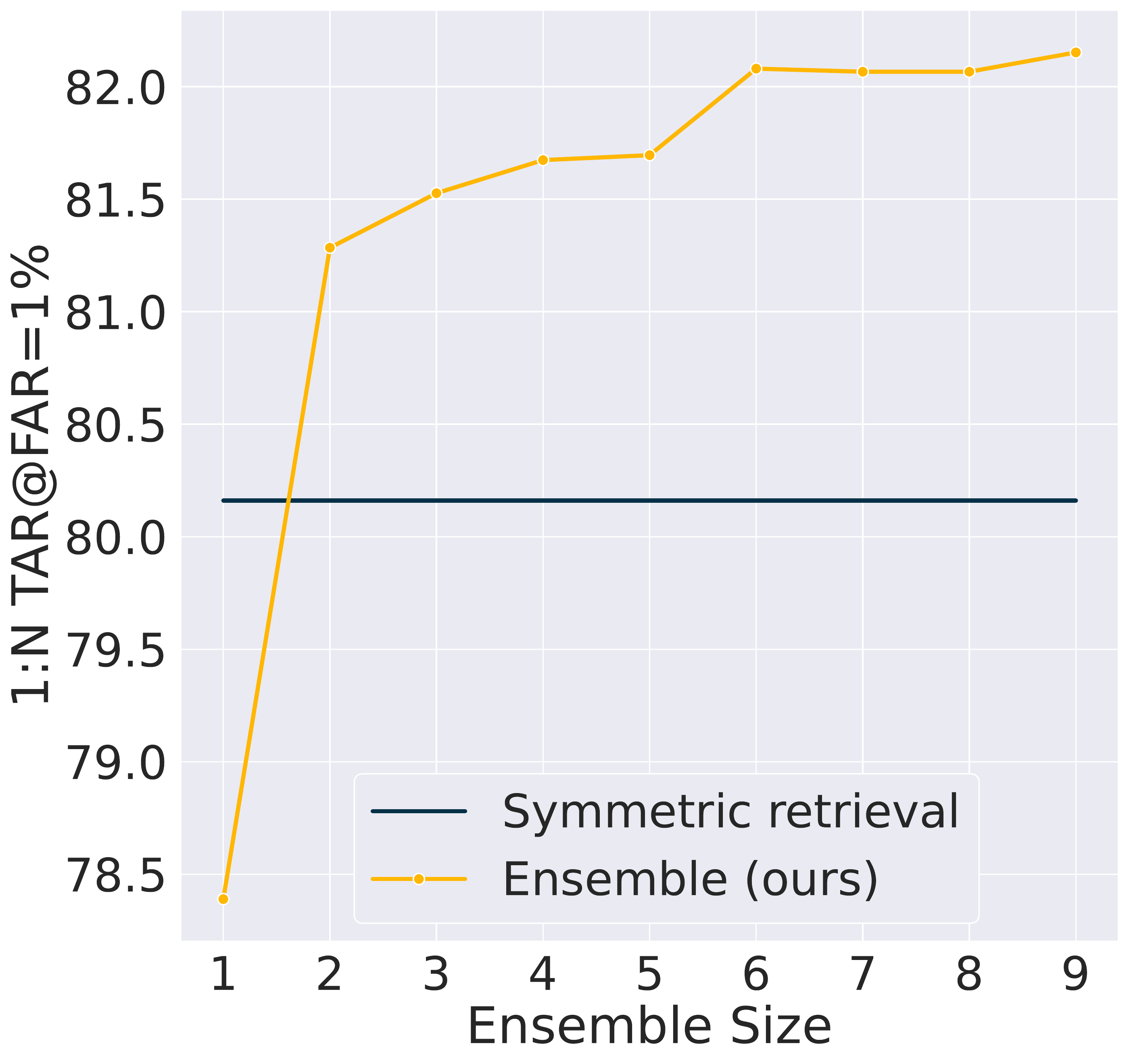}
         \caption{ResNet100 (1:N)}
         \label{fig:ensemble_size_r100_recog_1}
     \end{subfigure}
     \begin{subfigure}[b]{0.241\textwidth}
         \centering
         \includegraphics[width=0.88\textwidth]{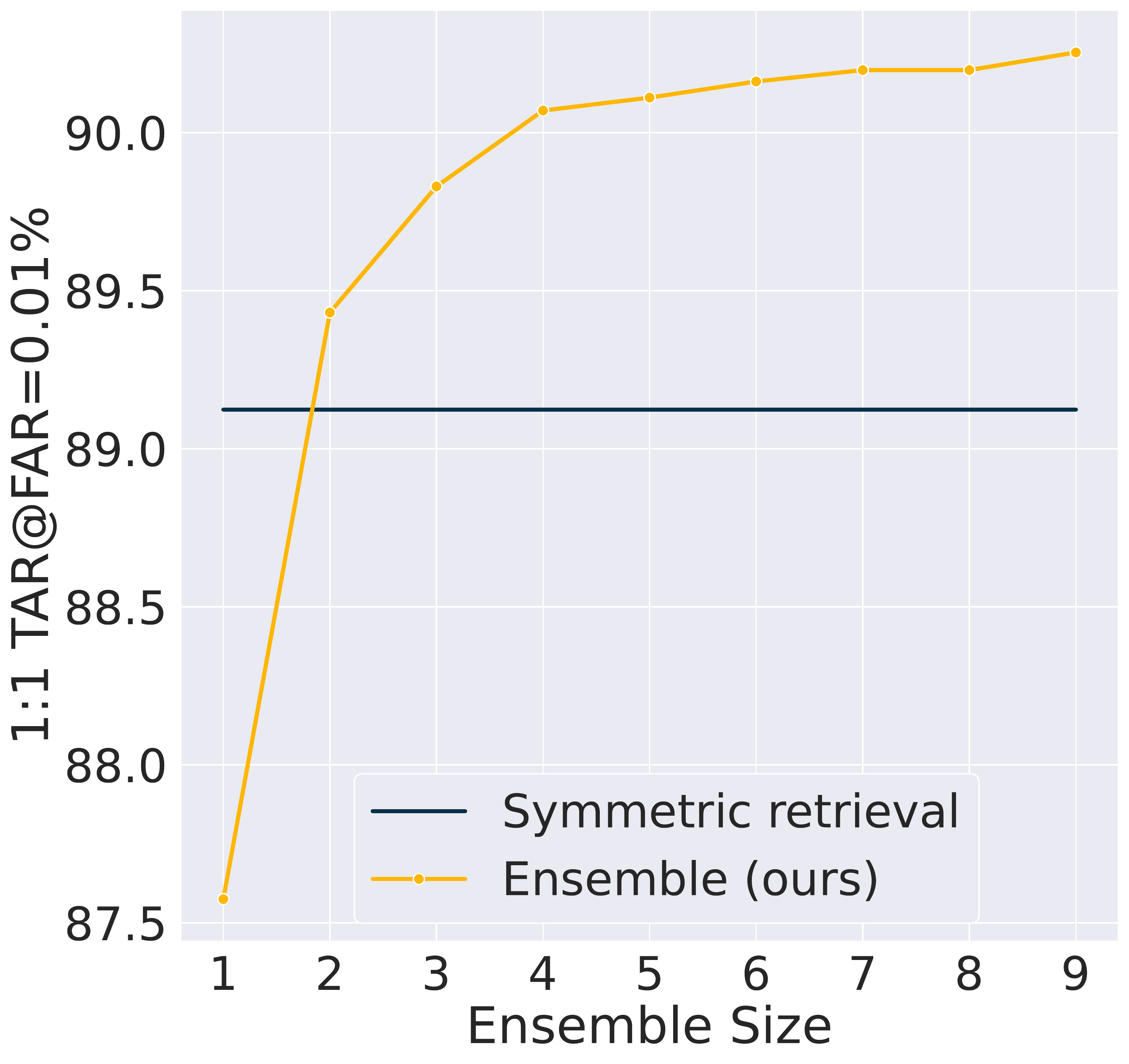}
         \caption{ResNet100 (1:1)}
         \label{fig:ensemble_size_r18_recog_1}
     \end{subfigure}
     \begin{subfigure}[b]{0.24\textwidth}
         \centering
         \includegraphics[width=0.88\textwidth]{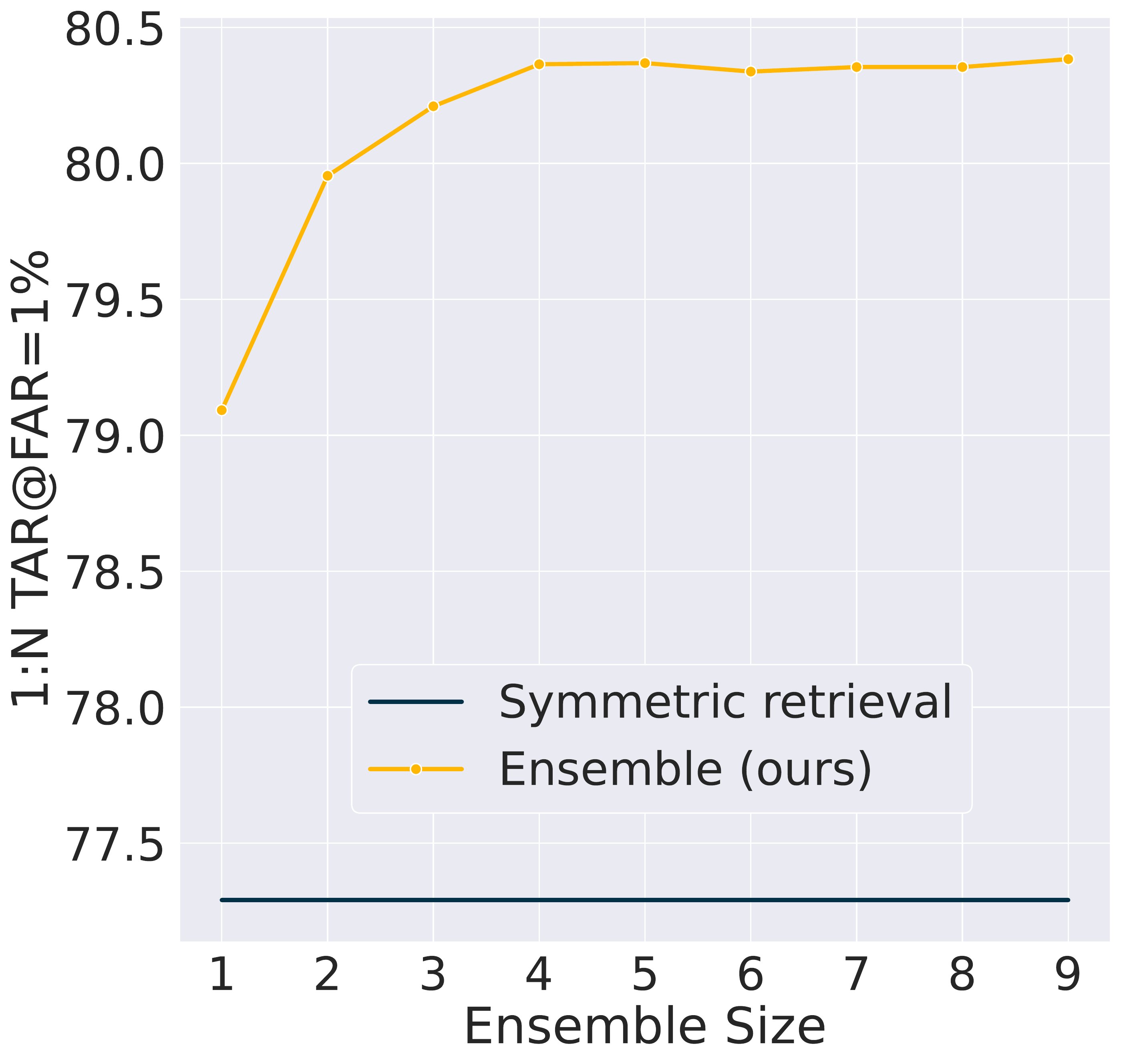}
         \caption{MBF (1:N)}
         \label{fig:ensemble_size_mbf_recog_1}
     \end{subfigure}
     \begin{subfigure}[b]{0.241\textwidth}
         \centering
         \includegraphics[width=0.88\textwidth]{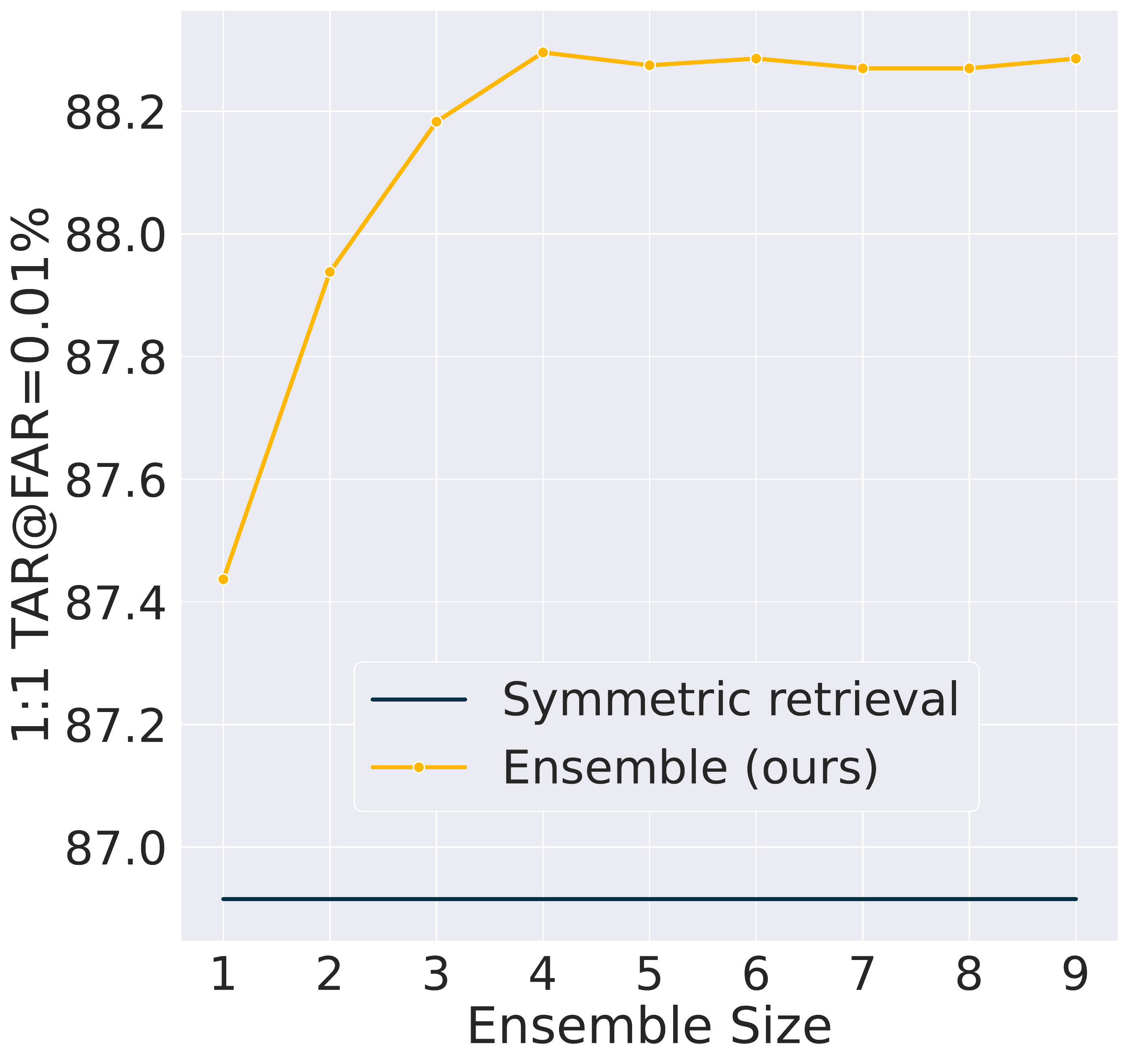}
         \caption{MBF (1:1)}
         \label{fig:ensemble_size_mbf_recog_1}
     \end{subfigure}
        \caption{1:1 verification and 1:N search accuracy vs. increasing ensemble sizes for models based on ResNet100 and MobileFaceNet.} 
    \vskip -0.1in
    \label{fig:ensemble_size}
\end{figure*}
\section{Experiments}
\label{sec:experiments}

We report experiments in two domains of image retrieval, \ie, face recognition and product retrieval.
We use face recognition as a particular case of image retrieval to evaluate and compare our method in various scenarios.
We then report results of the most interesting experiments on the product retrieval task.

\subsection{Face recognition experimental setup}
We use three common network architectures used in the face recognition domain: ResNet18, ResNet100~\cite{he2016deep} and MobileFaceNet (MBF)~\cite{chen2019r3}.
Each architecture was trained ten times on the VggFace2 dataset~\cite{Cao2018VGGFace2AD}, creating a total of 30 models. 
For evaluation, we follow Shen~\etal~\cite{shen2020towards} and utilize the widely used IJB-C benchmark~\cite{maze2018iarpa}.
We adopt the two standard testing protocols for face recognition, namely, 1:1 verification and 1:N open-set search.
In 1:1 face verification the algorithm decides whether a pair of templates belongs to the same person, where a template contains one or multiple images of a single person.
In 1:N open-set search, each query template is compared to a gallery of templates.
The algorithm then decides if and which gallery template matches the query template.
For both tasks, the evaluation metric is true acceptance rate (TAR) at a specific false acceptance rate (FAR).
For 1:1 verification we present TAR@FAR=0.01\% results, and for 1:N search we present TAR@FAR=1\%.

\subsection{Accuracy gains by ensemble size}
Fig.~\ref{fig:basline_comp} and Table~\ref{tab:basline_comp} provide results of our approach and the methods it builds upon.
We first present results for the symmetric retrieval setting ($\phi_g=\phi_q$), \ie, the standard use case.
We then present results for the unified embedding space approach~\cite{wang2020unified} (Unified) and the M2M approach described in~\ref{sec:m2m}.
Note that M2M corresponds to an ensemble of size one.
The results suggest that there is no significant difference between the two approaches, indicating that the M2M approach is a valid variant of the Unified approach for us to build upon.
Finally we combine multiple M2M models and show the significant accuracy gains achieved by the ensemble approach.
The ensembles of variable sizes were created incrementally by adding a single model each time. \Ie, ensemble of size $M+1$ contains exactly the same set of M2M models as for the ensemble of size $M$ with the addition of one new M2M model.
Note that knowledge distillation based approaches~\cite{duggal2021compatibility, Wu_2022_CVPR, Budnik_2021_CVPR, shen2020towards}, are upper-bounded by the performance of the stronger model (out of $\phi_g$ and $\phi_q$) which in Table~\ref{tab:basline_comp} corresponds to the result of symmetric retrieval.
Fig.~\ref{fig:ensemble_size} shows that the trend of improved accuracy as a function of increased ensemble size is preserved across different architectures.
Interestingly, in some cases the asymmetric transformation based approaches that do not use ensembles (Unified, M2M) receive a slightly better accuracy than the symmetric counterpart.
This phenomenon was also reported by Wang~\etal~\cite{wang2020unified} and might happen because of the additional transformation model parameters or by implicit regularization.



\subsection{Diversity of ensemble components}
\label{sec:diversity}
In this section we provide insights on how the diversity of the ensemble components impacts accuracy.
We consider three levels of component diversity:
\begin{enumerate}
    \item \textbf{Diversity of transformation models (D-T):}
    We use a single gallery model $\phi_g$ and train different transformation models mapping embedding vectors computed by $\phi_g$ to the query model's embedding space.
    \item \textbf{D-T + diversity of gallery models (D-TG):}
    We use $N$ different gallery models $\{\phi_g^i\}^{N}_{i=1}$ and train one transformation model per gallery model.
    Hence, both gallery models and transformation models are diverse. 
    \item \textbf{D-TG + diversity of gallery model architectures (D-TGA):}
    We further increase the diversity by allowing the gallery models to have different architectures.
\end{enumerate}
\begin{figure*}
    \begin{subfigure}[b]{0.49\textwidth}
         \centering
         \includegraphics[height=0.43\textwidth]{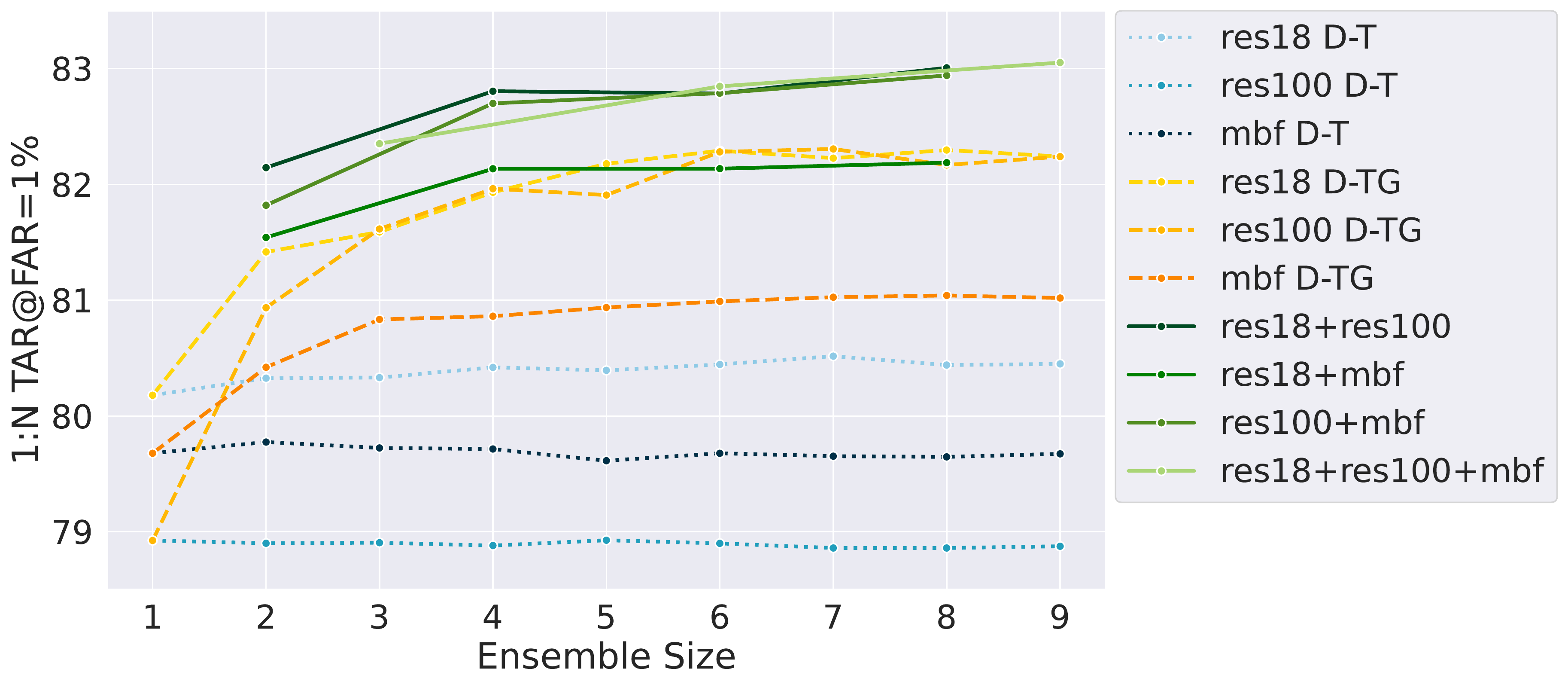}
         \caption{1:N TAR@FAR=1\%}
         \label{fig:diverse_ensemble_plot_rec}
     \end{subfigure}
     \begin{subfigure}[b]{0.49\textwidth}
         \centering
         \includegraphics[height=0.43\textwidth]{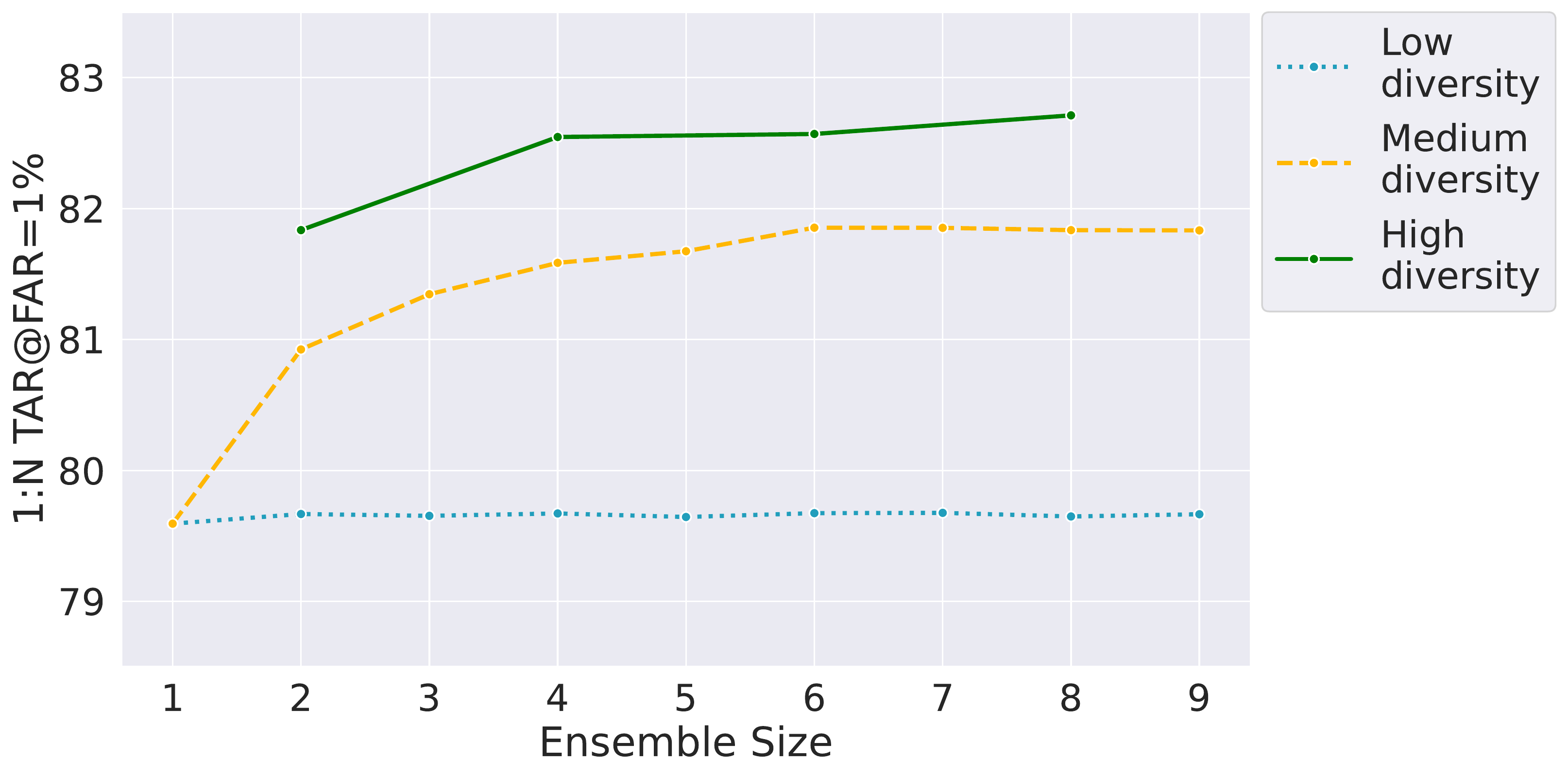}
         \caption{1:N TAR@FAR=1\% averaged by diversity level}
         \label{fig:diverse_ensemble_plot_ver}
     \end{subfigure}
    \caption{\textbf{Diversity of ensemble components.} 
    Dotted, dashed and solid lines correspond to D-T, D-TG and D-TGA results, respectively.
    For D-TGA the number of gallery models trained with each architecture is the same for each ensemble size, 
    \eg, res18+res100 of ensemble size 6 means three ResNet18 models and three ResNet100 models.
    (a) Shows the TAR for each ensemble type.
    (b) Shows the TAR for each diversity level by averaging the results of all ensemble types with the same diversity level and same ensemble size.
    The res18+res100+mbf D-TGA version was not used during averaging since this version ensemble sizes do not match the other D-TGA versions.
    }
    \vskip -0.1in
    \label{fig:diverse_ensemble_plot}
\end{figure*}

Fig.~\ref{fig:diverse_ensemble_plot} shows the accuracy for various ensembles with different levels of component diversity.
In all configurations the same ResNet18-based query model is used.
The results imply that component diversity plays an important role in the ensemble's performance.
Ensembles with high diversity consistently produce higher accuracy rates for the same number of transformation models than their less diverse counterparts.
Even when the ensemble is comprised of only three gallery models with different architectures (1xResNet18, 1xResNet100 and 1xMBF), we observe better results than using an ensemble of eight gallery models with the same ResNet18 architecture.
This result suggests that striving towards diverse ensembles is more beneficial than increasing the number of models in the ensemble.
We provide the following hypotheses to explain this phenomenon:
(1) each gallery model learns a different representation of the observed data~\cite{Wang2018TowardsUL, kornblith2019similarity, Mehrer2020IndividualDA} by focusing on slightly different features, 
(2) increased diversity in the gallery models results in more diverse representations, and
(3) combining diverse representations of the data leads to better generalization that is translated to better performance.
In other words, our method allows to capture more aspects of the data and combine them effectively.

\subsection{Comparison to other ensemble alternatives}
\label{sec:alternative}
To further validate our ensemble design choices, we present alternative approaches to our proposed method of using independently trained transformation models, one for each gallery model, and averaging the transformed embeddings.
Instead of the above, we learn a single combined transformation model that takes embeddings from all gallery models as inputs, and outputs a transformed embedding in the query model's embedding space.
Since creating a single combined transformation model is not straightforward, we propose several variants:
\begin{table}[t]
    \centering
    \begin{tabular}{l  c}
        \toprule
        Ensemble version & 1:N TAR@FAR=1\% \\
        \midrule
        End-to-end averaging & 81.195\% \\
        Weighted end-to-end averaging & 81.268\% \\
        Concatenation & 81.410\% \\
        Ours  & \textbf{81.932\%} \\
        \bottomrule
    \end{tabular}
    \caption{\textbf{Comparison to ensemble alternatives.}
    1:N search TAR@FAR=1\% results for each of the alternative ensemble variants.
    All experiments were conducted using the same query and same four gallery models.
    All models are based on ResNet18.}
    \vskip -0.1in
    \label{tab:table_alternative_ensemble}
\end{table}
\begin{figure}[t]
    \centering
    \includegraphics[width=0.42\textwidth]{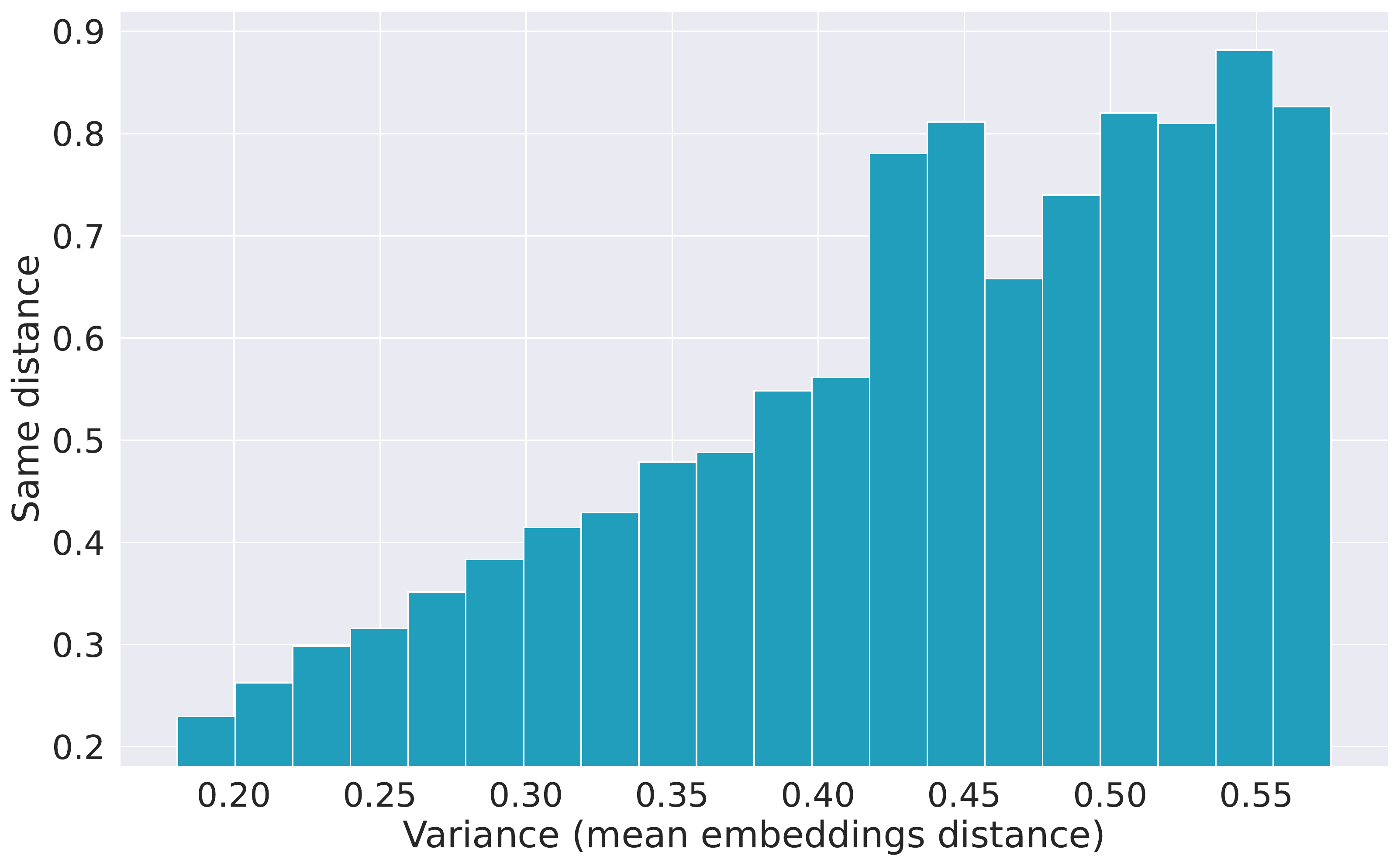}
    \vskip -0.1in
    \caption{\textbf{Same-distance vs. variance.}
    Each bar in the plot represents a range of variance levels (x-axis).
    The y-axis corresponds to the mean same-distance (\ie, the distance between the query embedding to a matching gallery embedding) of all matches inside the range.
    The bars' width is $0.248$ corresponding to 20 bins spanning from the lowest variance to the highest variance observed.}
    \vskip -0.1in
    \label{fig:same_dist_vs_var}
    
\end{figure}
\begin{figure*}[t]
    \centering
    \begin{subfigure}[b]{0.48\textwidth}
    \centering
    \includegraphics[width=0.98\textwidth]{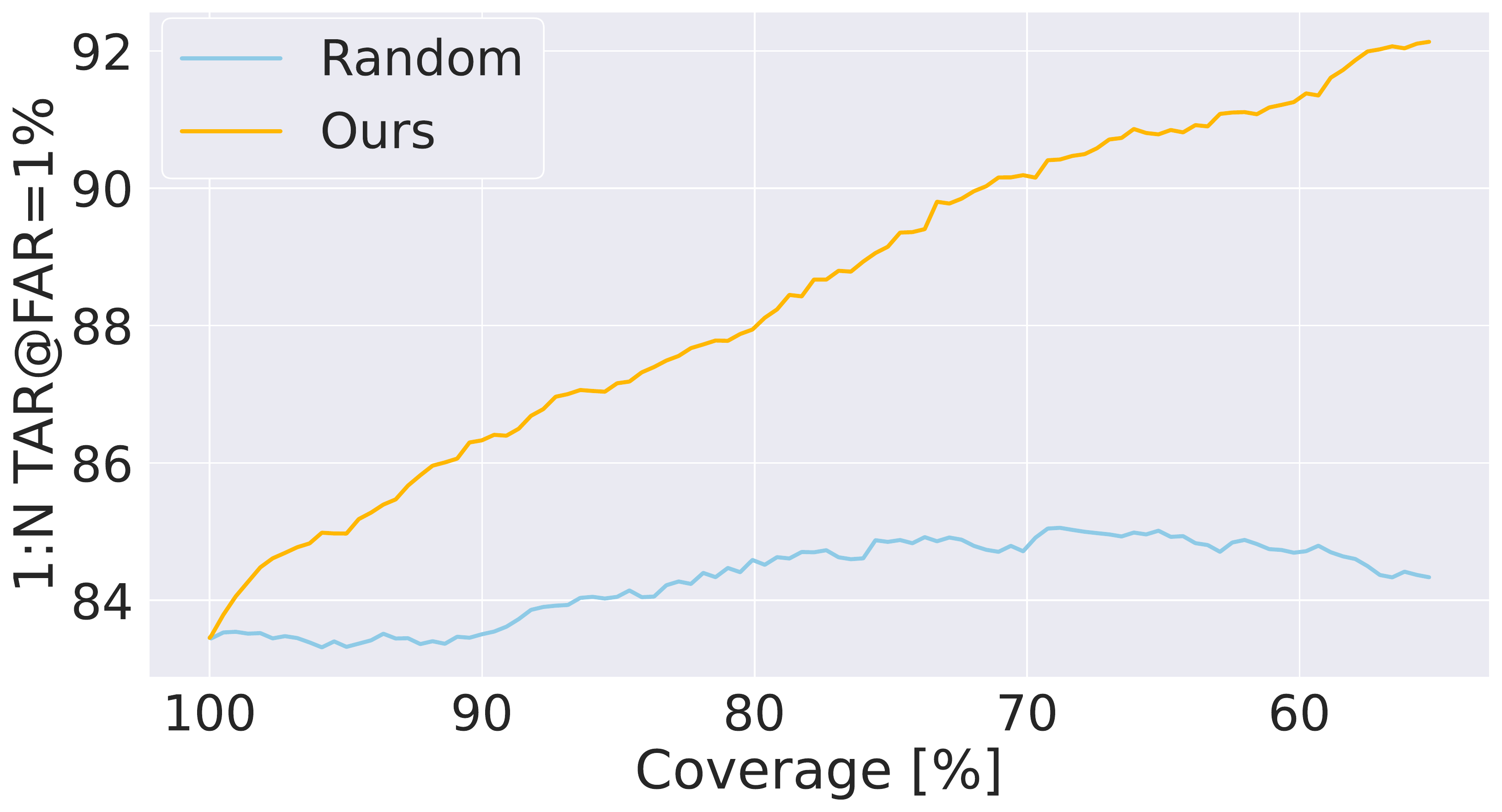}
    \caption{IJC-B 1:N TAR@FAR=1\% vs. coverage.}
    \label{fig:risk_cov_ijcb}
    \end{subfigure}
    \begin{subfigure}[b]{0.48\textwidth}
    \centering
    \includegraphics[width=0.98\textwidth]{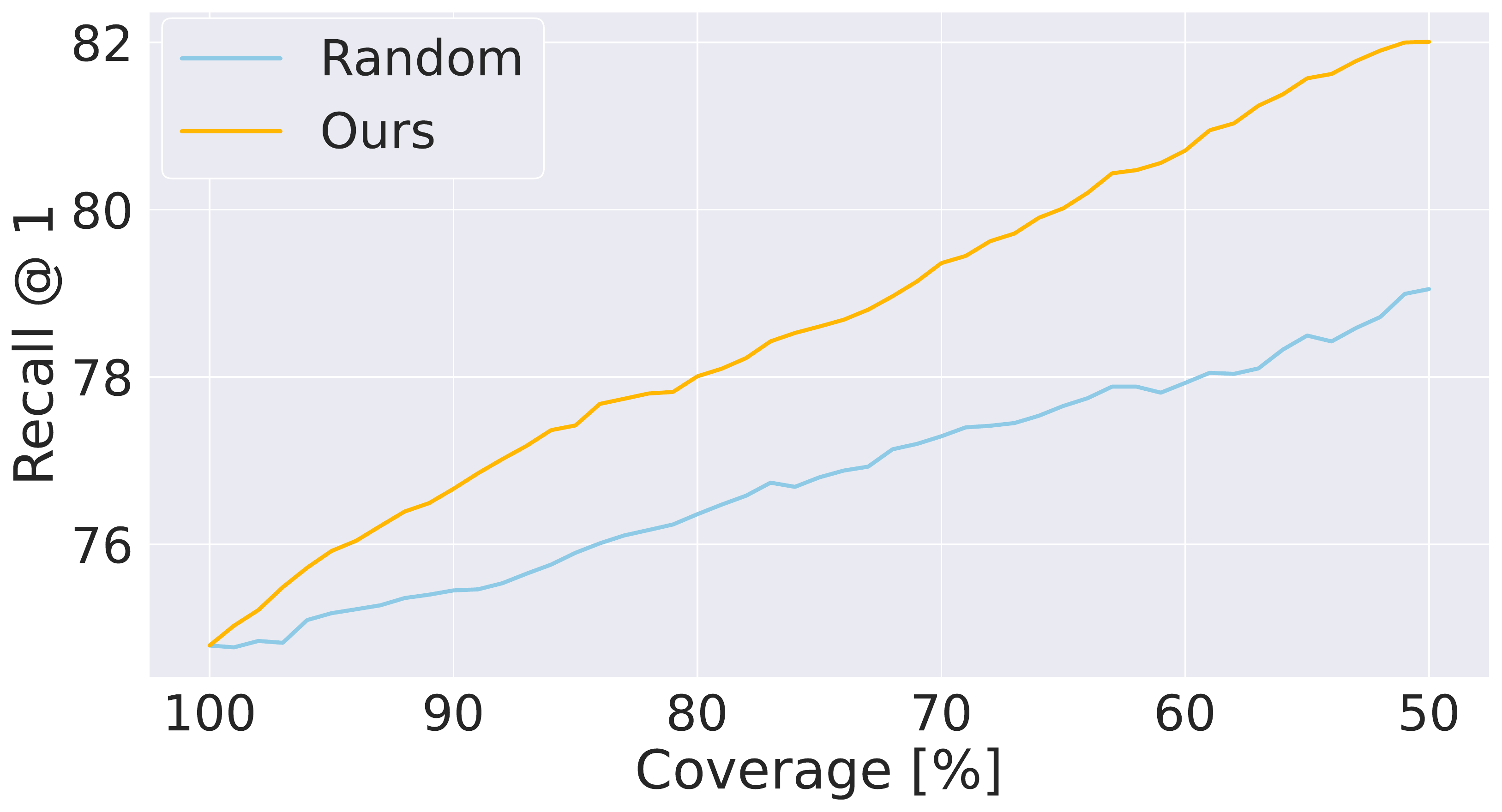}
    \caption{SOP Recall@1 vs. coverage.}
    \label{fig:risk_cov_sop}
    \end{subfigure}
    \vskip -0.05in
    \caption{\textbf{Risk-coverage curve.} We show the accuracy increase (risk) gained by filtering out (coverage) gallery embeddings using two protocols for both the IJC-B (a) and the SOP (b) datasets.
    Random: removing random embeddings (as a na\"ive baseline).
    Ours: rejecting embeddings whose variance is above a gradually decreasing threshold. 
    Note that, for Recall@1, smaller gallery sets are easier for matching, therefore, \textit{Random} shows improvement, but not to the same extent as ours.
    } 
    \vskip -0.2in
    \label{fig:risk_coverage}
\end{figure*}
\begin{enumerate}
    \item \textbf{End-to-end averaging:} We train a transformation model from each gallery model to the query embedding space simultaneously.
    Specifically, instead of training each transformation model independently, we train all transformation models together and perform the averaging during training.
    \item \textbf{Weighted end-to-end averaging:}  In a similar manner, we jointly train all transformation models except that we use weighted averaging instead of simple averaging.
    Each transformation model outputs a single scalar value, in addition to the embedding output, that is used for weighted averaging.
    \item \textbf{Concatenation:} All gallery embeddings are first concatenated and
    inserted into a linear layer that reduces the dimensions.
    The output of the linear layer is then transformed by a single transformation model to the embedding space of the query model.
\end{enumerate}
In this experiment, for all variants, and for our method, we used the same five pre-trained ResNet18 models. 
One for the query model, and four for the gallery models.

Table~\ref{tab:table_alternative_ensemble} presents the comparison between our ensemble approach and the combined transformation variants.
The results indicate that our approach performers considerably better then the other variants.
Furthermore, this implies that explicitly optimizing the ensemble performs worse than averaging independently trained models.

\subsection{Uncertainty}
It was previously observed that ensembles of models can be used to evaluate the uncertainty of model predictions\footnote{This type of uncertainty is typically known as \emph{epistemic} uncertainty or \emph{model} uncertainty.}~\cite{lakshminarayanan2016simple}. 
Typically, uncertainty is evaluated by calculating the variance between the predictions of the different models in the ensemble.
Thus, if the predictions of the different models are inconsistent, the uncertainty rises, and the models' predictions are considered less reliable.
In this section, we analyze whether the ensemble proposed in our approach can be used to reliably measure uncertainty.
To evaluate the model uncertainty for a given gallery image, we measure the variance of the transformed embeddings of the gallery models in the query embedding space.
The variance is calculated by measuring the mean distance between every pair of embeddings.

When querying, we consider the gallery embedding as \textit{easy to retrieve} if the distance between the gallery embedding to the query embedding of the same class is low (\emph{same-distance}). 
Fig.~\ref{fig:same_dist_vs_var} suggests that the same-distance and the measured variance between the transformed gallery embeddings are indeed correlated.
This result further implies that, upon registering a new gallery image, the variance can be used as a quality-score for accepting new gallery images, \ie, if the variance is larger than some threshold the image will be rejected.
In other words, if the gallery models' embeddings are far from one another, then the gallery embedding might not be easy to retrieve, thus we may choose to re-register with a new gallery image.

We follow Geifman~\etal~\cite{geifman2017selective} and measure the quality of uncertainty estimation by reporting the risk-coverage curve.
The curve represents the accuracy we can achieve (risk) as a function of the percent of gallery images we register (coverage).
The curve is constructed as following: as we decrease the threshold for uncertainty, more images are rejected.
For every point on the curve we provide the accuracy of the system given an uncertainty threshold.
The risk-coverage curve is presented in Fig.~\ref{fig:risk_cov_ijcb},
and demonstrates that even rejecting as few as 10\% of the gallery embeddings, reduces the error by 17.4\%, while a 20\% rejecting rate leads to a 27.2\% error reduction.
Fig.~\ref{fig:risk_cov_ijcb} shows that our filtering method is consistently better than na\"ively removing gallery images, meaning that the improved accuracy is not the sheer result of reduction in the number of gallery images, rather in reduction of images that are prone to error.   
This suggest that the variance of gallery embeddings could be used as an embedding quality metric.
In the supplementary, we provide examples for gallery images with varying uncertainty levels.  


\subsection{Compatible query model update}
\label{sec:query_model_update}
In this section we evaluate the scenario where an improved query model was developed, replacing an older one and backfilling is not possible.
This simulates a system that does not retain data (except of training data).
We simulate the scenario as follows:
\begin{enumerate}
    \item \textbf{Before update:} The system is composed of $N$ gallery models and one query model.
    All models were trained on 50\% of the VggFace2 dataset, that represents the available training dataset at some point during the lifespan of the system.
    At this stage the gallery set is mapped to the embedding spaces of all gallery models ($\{\phi^{50\%}_{g,i}\}_{i=1}^N$) and the gallery images are discarded.
    We transform the embeddings from the embedding spaces of $\{\phi^{50\%}_{g,i}\}_{i=1}^N$ to the embedding space of the query model ($\phi^{50\%}_q$), using our ensemble approach.
    \item \textbf{After update:}
    At some point additional training data was made available\footnote{Additional training data can be acquired by conducting a dedicated data collection effort for example.} and was used to train an improved query model.
    We simulate this by training a new query model on 100\% of the VggFace2 dataset ($\phi^{100\%}_q$).
    This time the corresponding embedding spaces of $\{\phi^{50\%}_{g,i}\}_{i=1}^N$ are transformed to the embedding space of $\phi^{100\%}_q$ using new transformation models.
\end{enumerate}

The above scenario represents an extreme case, where the performance gap between the old models and the newer one (trained on 50\% and 100\% of the data respectively) is significant.
This can be seen in Table~\ref{tab:query_model_update} where the 1:N search TAR@FAR=1\% of $\phi^{50\%}_q$ is 71.01\% vs. 79.93\% of $\phi^{100\%}_q$.

Table~\ref{tab:query_model_update} demonstrates the performance gain of updating the query model.
Interestingly, the performance increases drastically compared to the gap before the update (for ensemble size of four, the 1:N search TAR@FAR=1\% increases from 73.98\% to 78.96\%), despite using only low-accuracy-models for gallery indexing.
Furthermore, Table~\ref{tab:query_model_update} shows that a larger ensemble size corresponds with improved performance even when the query model is much stronger than the gallery models used for indexing.

\begin{table}[]
    \centering
    \vspace{3pt}
    \small
    \begin{tabular}{l c c}
        \toprule
        \quad\quad\quad\quad\quad\quad\quad\quad\quad\quad\quad\quad\quad\quad\quad & 1:N TAR & 1:1 TAR \\
        \midrule
        \multicolumn{3}{c}{\footnotesize{Independent models in the symmetric setting}}\\ \midrule
        Gallery models: $\{\phi^{50\%}_{g,i}\}_{i=1}^4$ & 71.96\tiny{$\pm 0.3$} & 83.24\tiny{$\pm 0.17$} \\
        Query model: $\phi^{50\%}_q$ & 71.01 & 83.00 \\
        Query model: $\phi^{100\%}_q$ \textdagger  & 79.93 & 88.28 \\
        \midrule
        \multicolumn{3}{c}{\footnotesize{Ensemble of size 2 ($\{\phi^{50\%}_{g,i}\}_{i=1}^2$)}} \\ \midrule
        Before update ($\phi^{50\%}_q$) & 72.90 & 83.78 \\
        After update ($\phi^{100\%}_q$) & 78.21 & 86.94 \\
        \midrule \multicolumn{3}{c}{\footnotesize{Ensemble of size 4 ($\{\phi^{50\%}_{g,i}\}_{i=1}^4$)}} \\ \midrule
        Before update ($\phi^{50\%}_q$) & 73.98 & 84.62 \\
        After update ($\phi^{100\%}_q$) & \textbf{78.96} & \textbf{87.39} \\
        \bottomrule
    \end{tabular}
    \caption{\textbf{Compatible query model update.} 
    First three rows show the accuracy of the standard symmetric setting of all models used in the experiment (the accuracy of the gallery models is presented as the average of four models).  
    ``Before update'', shows the results where querying is done by the old query model, $\phi^{50\%}_q$.
    ``After update'', shows the results for using the updated query model, $\phi^{100\%}_q$, for querying.
    In both, before and after update, the same gallery models are used for indexing. 
    \textdagger: Note that in the no-backfilling scenario using $\phi^{100\%}_q$ in a symmetric setting is not possible.
    }
    \vskip -0.1in
    \label{tab:query_model_update}
\end{table}

\begin{figure}[t]
    \centering
    \includegraphics[width=0.88\linewidth]{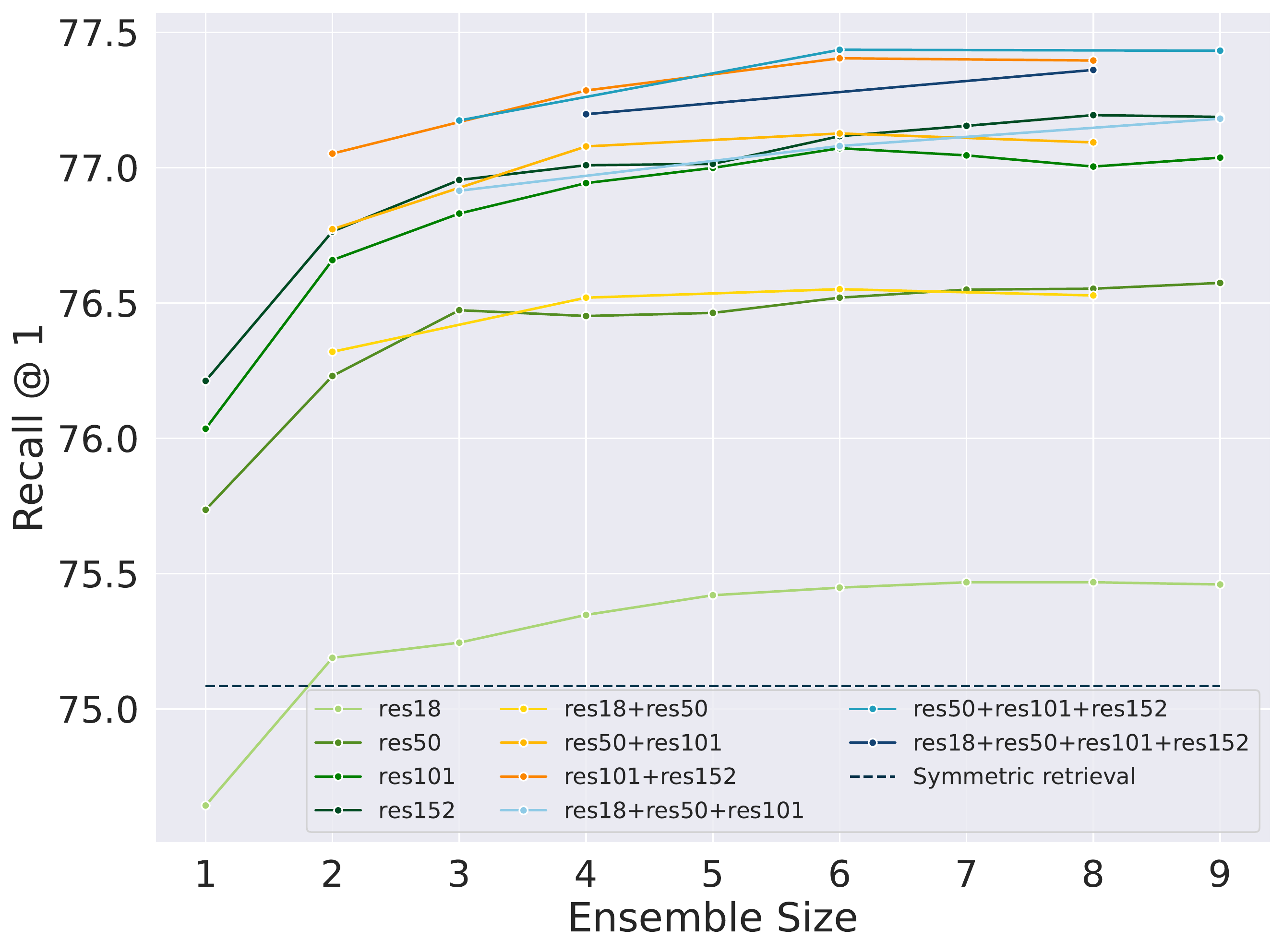}
    \caption{\textbf{Recall@1 vs. ensemble size.} ``Symmetric retrieval'' denotes the accuracy of the ResNet18 query model. 
    }
    \vskip -0.12in
    \label{fig:diverse_sop_ensemble_plot}
\end{figure}
\definecolor{ar}{RGB}{255, 0, 0}
\definecolor{ag}{RGB}{63, 192, 86}
\definecolor{ab}{RGB}{25 , 76, 230}
\definecolor{ao}{RGB}{253 , 119, 6}

\begin{figure}
\centering
\begin{subfigure}[b]{0.98\linewidth}
        \centering
        \includegraphics[width=0.99\linewidth]{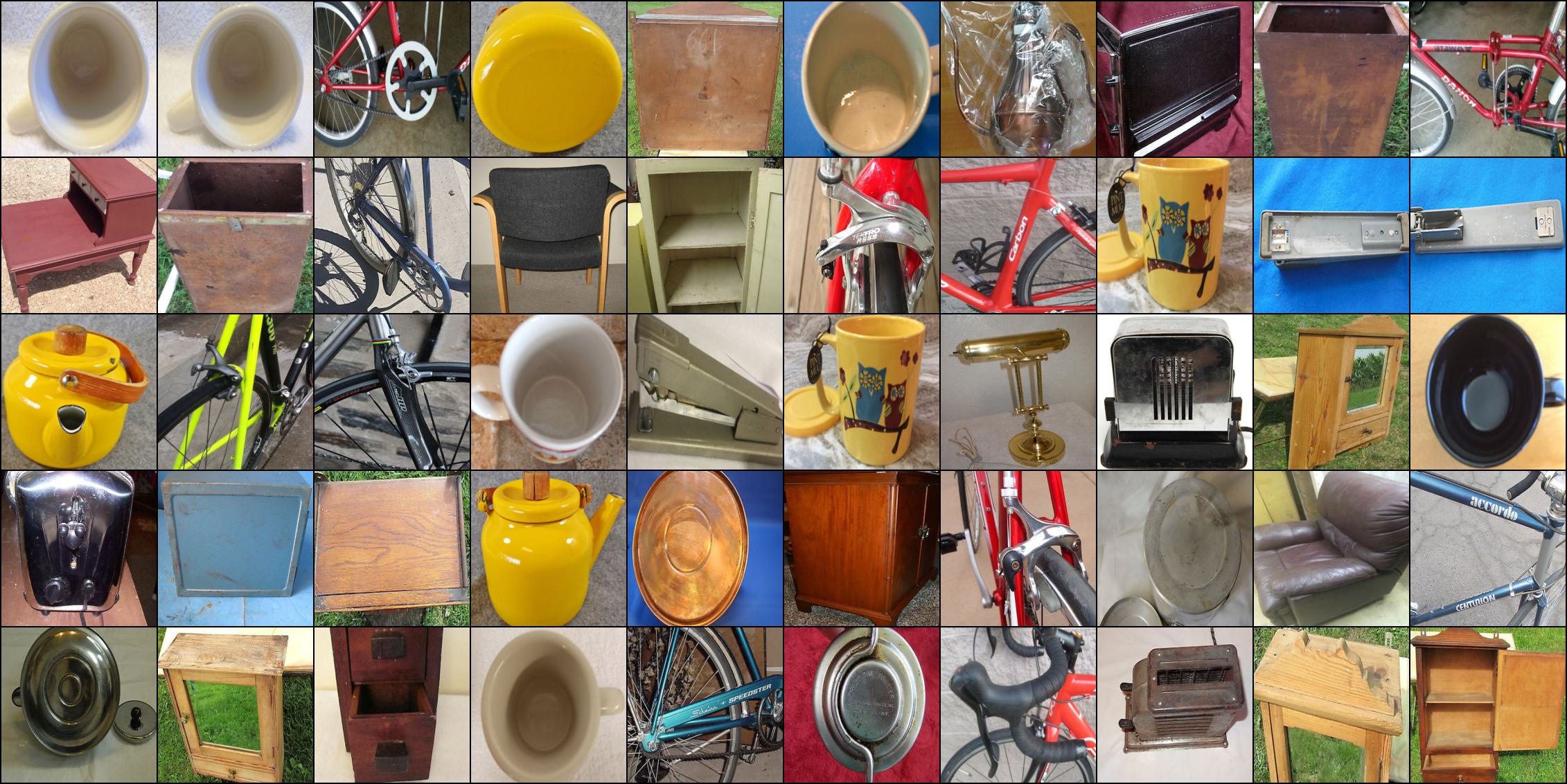}
        \caption{50 images with the lowest uncertainty.}
        \label{fig:ddd}
\end{subfigure}
\begin{subfigure}[b]{0.98\linewidth}
        \centering
        \includegraphics[width=0.99\linewidth]{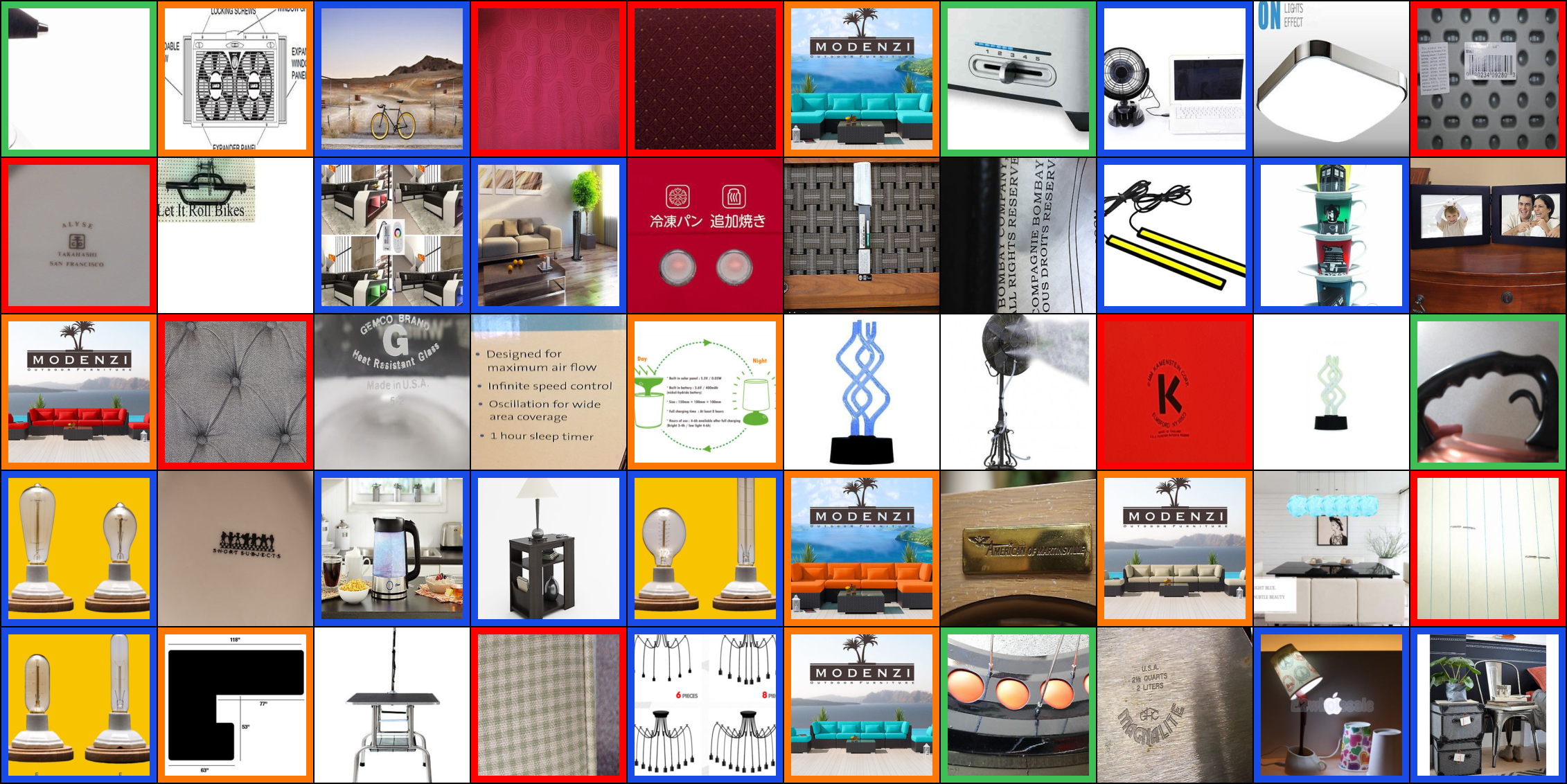}
        \caption{50 images with the highest uncertainty.}
        \label{fig:ccc}
\end{subfigure}
\centering

\caption{\textbf{Examples of SOP gallery images.} (a) and (b) show the images with the lowest and highest variance values respectively.
Note the various factors resulting in high ambiguity within the ensemble: \textcolor{ar}{extreme zoom}, \textcolor{ag}{partial views}, \textcolor{ab}{multiple objects}, \textcolor{ao}{non-natural images} (\eg, sketched, rendered). [can be zoomed-in].}
\vskip -0.1in
\label{fig:sop_images}
\end{figure}
\subsection{Product retrieval experiments}
To further demonstrate the generality of our approach we conducted experiments on the Stanford online products (SOP) dataset~\cite{Song2015DeepML}.
The dataset contains 22.6K classes with 120K product images, where 11.3K classes (59.5K images) are used for training and the remaining 11.3K classes (60.5K images) are used for testing.
We followed the generic protocol proposed by~\cite{Zhai2018ClassificationIA} for training query and gallery models. 
We used the Recall@1~\cite{Jgou2011ProductQF} metric for evaluating the retrieval performance.
In all the experiments we used the same ResNet18 model for querying.
Fig.~\ref{fig:diverse_sop_ensemble_plot} shows that the benefits from using an ensemble generalize to the domain of product retrieval.
Additionally, Fig.~\ref{fig:diverse_sop_ensemble_plot} repeats the previously observed trend where diverse ensembles are generally preferable. 
Fig.~\ref{fig:risk_cov_sop} demonstrates the benefit of filtering out gallery embeddings with high uncertainty in this domain.
To conform with the previous settings for which we calculated the risk-coverage curve, for each test label, we used a single image as query and a single image for the gallery set.
Fig.~\ref{fig:sop_images} shows the images with lowest and highest variance in the gallery set.
Low variance images generally include a single identifiable object, while high variance images suffer from multiple ambiguity factors.

\section{Conclusions}
\label{sec:conclusions}
We propose a novel embedding transformation based ensemble framework for asymmetric image retrieval.
We show that embedding transformations can be leveraged for creating a non-trivial ensemble of diverse gallery models, significantly increasing the retrieval accuracy without increasing the computational cost of querying.
We compared several methods for combining multiple embedding spaces and found that training the transformation models independently lead to the best performance.
Additionally, we utilize the diversity between multiple transformed embeddings to estimate the uncertainty of gallery images.
We propose to reject gallery images based on their uncertainty to further improve our system's accuracy.





\newpage
{\small
\bibliographystyle{ieee_fullname}
\bibliography{egbib}
}

\end{document}